\documentclass{l4dc2024}

\title[Growing Q-Networks]{Growing Q-Networks:\\Solving Continuous Control Tasks with Adaptive Control Resolution}
\usepackage{times}

\usepackage{hyperref}
\usepackage{url}

\usepackage{mathtools}
\usepackage{mleftright}
\usepackage{bm}
\usepackage{placeins}
\usepackage{siunitx}
\usepackage{wrapfig}
\usepackage{booktabs}

\usepackage{xcolor}

\newcommand{\sgn}{\operatorname{sgn}}

\author{%
 \Name{Tim Seyde} \Email{tseyde@mit.edu}\\
 \addr MIT CSAIL
 \AND
 \Name{Peter Werner} \Email{wernerpe@mit.edu}\\
 \addr MIT CSAIL%
 \AND
 \Name{Wilko Schwarting} \Email{wilko@isee.ai}\\
 \addr ISEE AI%
 \AND
 \Name{Markus Wulfmeier*} \Email{mwulfmeier@google.com}\\
 \addr Google DeepMind%
 \AND
 \Name{Daniela Rus*} \Email{rus@csail.mit.edu}\\
 \addr MIT CSAIL%
}

\begin{document}

\maketitle

\begin{abstract}%
 Recent reinforcement learning approaches have shown surprisingly strong capabilities of bang-bang policies for solving continuous control benchmarks. The underlying coarse action space discretizations often yield favourable exploration characteristics while final performance does not visibly suffer in the absence of action penalization in line with optimal control theory. In robotics applications, smooth control signals are commonly preferred to reduce system wear and energy efficiency, but action costs can be detrimental to exploration during early training. In this work, we aim to bridge this performance gap by growing discrete action spaces from coarse to fine control resolution, taking advantage of recent results in decoupled Q-learning to scale our approach to high-dimensional action spaces up to $dim(\mathcal{A})=38$. Our work indicates that an adaptive control resolution in combination with value decomposition yields simple critic-only algorithms that yield surprisingly strong performance on continuous control tasks.%
\end{abstract}


\section{Introduction}

Reinforcement learning for continuous control applications commonly leverages policies parameterized via continuous distributions. Recent works have shown surprisingly strong performance of discrete policies both in the actor-critic and critic-only setting~\citep{tang2020discretizing, tavakoli2021learning, seyde2021bang}.
While discrete critic-only methods promise simpler controller designs than their continuous actor-critic counterparts, applications such as robot control tend to favor smooth control signals to maintain stability and prevent system wear~\citep{Hodel2018}.
It has previously been noted that coarse action discretization can provide exploration benefits early during training~\citep{czarnecki2018mix, farquhar2020growing}, while converged policies should increasingly prioritize controller smoothness~\citep{bohez2019value}.
Our work aims to bridge the gap between these two objectives while maintaining algorithm simplicity. We introduce Growing Q-Networks (GQN), a simple discrete critic-only agent that combines the scalability benefits of fully decoupled Q-learning~\citep{seyde2022solving} with the exploration benefits of dynamic control resolution~\citep{czarnecki2018mix, farquhar2020growing}.
Introducing an adaptive action masking mechanism into a value-decomposed Q-Network, the agent can autonomously decide when to increase control resolution. This approach enhances learning efficiency and balances the exploration-exploitation trade-off more effectively, improving convergence speed and solution smoothness.
The primary contributions of this paper are threefold: 
\begin{itemize}
\item \textbf{Framework for adaptive control resolution:} 
we adaptively grow control resolution from coarse to fine within decoupled Q-learning. This reconciles coarse exploration during early training with smooth control at convergence, while retaining scalability of decoupled control.
\item \textbf{Insights into scalability of discretized control:} our research provides valuable insights into overcoming exploration challenges in soft-contrained continuous control settings via simple discrete Q-learning methods, studying applicability in challenging control scenarios.
\item \textbf{Comprehensive experimental validation:} we validate the effectiveness of our GQN algorithm on a diverse set of continuous control tasks, highlighting benefits of adaptive control resolution over static DQN variations as well as recent continuous actor-critic methods.
\end{itemize}

The remainder of the paper is organized as follows: Section 2 reviews related work, Section 3 introduces preliminaries, Section 4 details the proposed GQN methodology, Section 4 presents experimental results, and Section 5 concludes with a discussion on future research directions.

\section{Related Works}
In the following, we discuss several key related works grouped by their primary research thrust.
\paragraph{Discretized Control}
Learning continuous control tasks typically relies on policies with continuous support, primarily Gaussians with diagonal covariance matrices~\citep{schulman2017proximal,haarnoja2018soft,abdolmaleki2018relative,hafner2020mastering, Wulfmeier2020a}.
Recent works have shown that competitive performance is often attainable via discrete policies~\citep{tavakoli2018action, neunert2020continuous, tang2020discretizing, seyde2022strength} with bang-bang control at the extreme~\citep{seyde2021bang}.
Bang-bang controllers have been extensively investigated in optimal control research~\citep{Sonneborn1964,Bellman1956,LaSalle1959,Maurer2005} as well as early works in reinforcement learning~\citep{Waltz1965,Lambert1970,Anderson1988}, while the extreme switching behavior was often observed to naturally emerge even under continuous policy distributions~\citep{Huang2019,Novati2019,Thuruthel2019}.
The direct application of discrete action-space algorithms then harbors potential benefits for reducing model complexity~\citep{metz2017discrete, sharma2017learning, tavakoli2021structural, watkins1992q}, although control resolution trade-offs and scalability may require computational overhead~\citep{van2020q}.

\paragraph{Scalability}
Scalability of Q-learning based approaches has been studied extensively in the context of mitigating coordination challenges and system non-stationarity~\citep{tan1993multi, claus1998dynamics, matignon2012independent, lauer2000algorithm, matignon2007hysteretic, foerster2017stabilising,busoniu2006decentralized, bohmer2019exploration}.
Exponential coupling can be avoided by information-sharing~\citep{schneider1999distributed, russell2003q, yang2018mean}, composition of local utility functions~\citep{sunehag2017value, rashid2018qmix, son2019qtran, wang2020dop, su2021value, peng2021facmac}, and considering different levels of interaction~\citep{guestrin2002coordinated, kok2006collaborative}. 
Centralization can further be facilitated via high degrees of parameter-sharing~\citep{gupta2017cooperative, bohmer2020deep, christianos2021scaling, van2017hybrid, chu2017parameter}).
Decoupled control via Q-learning was proposed for Atari~\citep{sharma2017learning} and extended to mixing across higher-order action subspaces~\citep{tavakoli2021learning}, with decoupled bang-bang control displaying strong performance on continuous control tasks~\citep{seyde2022solving}.
While coarse discretization can be beneficial for exploration, particularly in the presence of action penalties, they may also reduce steady-state performance. 
Conversely, fine discretization can exacerbate coordination challenges~\citep{seyde2022solving, ireland2024revalued}.
Here, we consider adapting the control resolution over the course of training to achieve the best of both worlds.

\paragraph{Expanding Action Spaces}
\cite{smith2023grow} presents an adaptive policy regularization approach that introduces soft constraints on feasible action regions, growing continuous regions linearly over the course of training with adjustments based on dynamics uncertainty. They focus on learning quadrupedal locomotion on hardware and expand locally around joint angles of a stable initial pose.
In discrete action spaces, one can instead leverage iterative resolution refinement.
\cite{czarnecki2018mix} considers DeepMind Lab navigation tasks~\citep{beattie2016deepmind} with a natively discrete action space that avoids reasoning about system dynamics stability. Their policy-based method formulates a mixture policy that is optimized under a distillation objective to facilitate knowledge transfer, adjusting the mixing weights via Population Based Training (PBT)~\citep{jaderberg2017population}.
Similarly, \cite{synnaeve2019growing} considers multi-agent coordination in StarCraft and adjusts spatial command resolution via PBT. 
\cite{farquhar2020growing} grow action resolution under a linear growth schedule, while showing limited application to simple continuous control tasks, as they enumerate the action space and do not consider decoupled optimization.
Beyond control applications, \cite{yang2023reinforcement} demonstrate adaptive mesh refinement strategies that reduce the errors in finite element simulations. Their refinement policy recursively adds finer elements, expanding the action space.
\paragraph{Constrained Optimization} Reward-optimal bang-bang policies may not be desirable for real-world applications as they can be less energy efficient and increase wear and tear on physical systems, e.g.~\cite{Hodel2018}. In the past, this behavior was generally avoided by employing penalty functions as soft constraints at the cost of potentially hindering exploration or enabling reward hacking~\cite{skalse2022defining}. The rewards and costs are automatically re-balanced to combat this issue in~\cite{bohez2019value}. Similarly, undesirable behaviors are avoided by automatically balancing soft chance constraints with the primary rewards in~\cite{roy2021direct}.
Here, we do not assume access to explicit penalty terms and efficiently learn controllers directly based on environment reward.

\section{Preliminaries}
We formulate the learning control problem as a Markov Decision Process (MDP) described by the tuple $\{\mathcal{S}, \mathcal{A}, \mathcal{T}, \mathcal{R}, \gamma\}$, where $\mathcal{S} \subset \mathbb{R}^{N}$ and $\mathcal{A} \subset \mathbb{R}^{M}$ denote the state and action space, respectively, $\mathcal{T}: \mathcal{S} \times \mathcal{A} \rightarrow \mathcal{S}$ the transition distribution, $\mathcal{R}: \mathcal{S} \times \mathcal{A} \rightarrow \mathbb{R}$ the reward function, and $\gamma \in [0, 1)$ the discount factor. 
Let $s_t$ and $a_t$ denote the state and action at time $t$, where actions are sampled from policy $\pi \mleft(a_t | s_t\mright)$. We define the discounted infinite horizon return as $G_t = \sum_{\tau=t}^{\infty}\gamma^{\tau-t} R\mleft(s_{\tau}, a_{\tau}\mright)$, where $s_{t+1} \sim \mathcal{T}\mleft(\cdot | s_t, a_t\mright)$ and $a_t \sim \pi\mleft(\cdot | s_t\mright)$.
Our objective is to learn the optimal policy that maximizes the expected infinite horizon return $\mathbb{E}[G_t]$ under unknown dynamics and reward mappings.
Conventional algorithms for continuous control settings leverage actor-critic designs with a continuous policy $\pi_{\phi}\mleft(a_t | s_t\mright)$ maximizing expected returns from a value estimator $Q_{\theta}\mleft(s_t, a_t\mright)$ or $V_{\theta}\mleft(s_t\mright)$.
Recent studies have shown strong results with simpler methods employing discretized actors~\citep{tang2020discretizing, seyde2021bang} or critic-only formulations~\citep{tavakoli2018action, tavakoli2021learning, seyde2022solving}.
Here, we focus on the light-weight critic-only setting and increase control resolution over the course of training to bridge the gap between discrete and continuous control.

\subsection{Deep Q-Networks}
We consider the general framework of Deep Q-Networks (DQN)~\citep{mnih2013playing}, where the state action value function $Q_{\theta}\mleft(s_t, a_t\mright)$ is represented by a neural network with parameters $\theta$. The parameters are updated to minimize the temporal-difference (TD) error, where we leverage several performance enhancements based on the Rainbow agent~\citep{hessel2018rainbow}.
These include target networks to improve stability in combination with double Q-learning to mitigate overestimation~\citep{mnih2015human,van2016deep}, prioritized experience replay (PER) to focus sampling on more informative transitions~\citep{schaul2015prioritized}, and multi-step returns to improve stability of Bellman backups~\citep{sutton2018reinforcement}. The resulting objective function is given by
\begin{align}
\label{eq:loss}
    \mathcal{L}(\theta) = \sum_{b=1}^{B}L_{\delta}\mleft(y_t - Q_{\theta}\mleft(s_t, a_t\mright)\mright),
\end{align}
where action evaluation employs the target $y_t=\sum_{j=0}^{n-1}\gamma^{j}r\mleft(s_{t+j}, a_{t+j}\mright) + \gamma^n Q_{\theta^-}\mleft(s_{t+n}, a^*_{t+n}\mright)$, action selection uses $a^*_{t+1} = \arg\max_a Q_{\theta}\mleft(s_{t+1}, a\mright)$, $L_{\delta}(\cdot)$ is the Huber loss and the batch size is $B$.

\subsection{Decoupled Q-Networks}
Traditional DQN-based agents enumerate the entire action space and do therefore not scale well to high dimensional control problems.
Decoupled representations address scalability issues by treating subsets of action dimensions as separate agents and coordinating joint behavior in expectation~\citep{sharma2017learning,sunehag2017value,rashid2018qmix,tavakoli2021learning,seyde2022solving}.
The Decoupled Q-Networks (DecQN) agent introduced in \cite{seyde2022solving} employs a full decomposition with the critic predicting univariate utilities for each action dimension $a^j$ conditioned on the global state $\bm{s}$.
The corresponding state-action value function is recovered as
\begin{align}
\label{eq:value_decomp}
    Q_{\theta}(\bm{s}_t, \bm{a}_t) = \sum_{j=1}^{M} \frac{Q_{\theta}^j(\bm{s}_t, a^j_t)}{M},
\end{align}
where the objective is analogous to Eq.~\ref{eq:loss}, enabling centralized training with decentralized execution.

\section{Growing Q-Networks}
\label{sec:gqn}
\begin{figure}[t!]
    \includegraphics[width=1.0\linewidth]{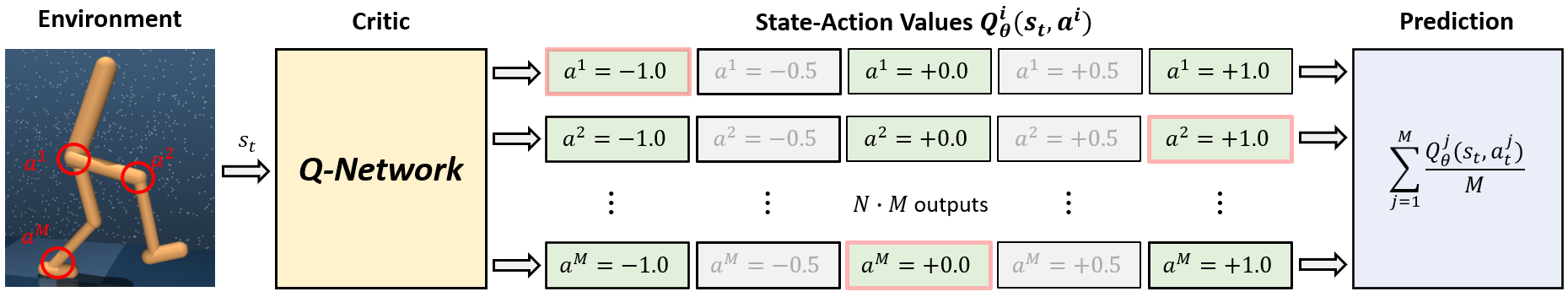}
\caption{Schematic of a GQN agent with decoupled $5$-bin discretization and $3$-bin active subspace. The available actions are highlighted in green while the masked actions are depicted in gray. The predicted state-action values $Q(\bm{s}, a^0, ..., a^M)$ are computed via linear composition of the univariate utilities $Q(\bm{s}, a^j)$ by selecting one action per dimension (red).}
\vspace{-0.6cm}
\label{fig:gqn_schematic}
\end{figure}%
Discrete control algorithms have demonstrated competitive performance on continuous control benchmarks~\citep{tang2020discretizing,tavakoli2018action,seyde2021bang}. One potential benefit of these methods is the intrinsic coarse exploration that can accelerate the generation of informative environment feedback. In robot control, we typically prefer smooth controllers at convergence to limit hardware stress. Our objective is to bridge the gap between coarse exploration capabilities and smooth control performance while retaining sample-efficient learning.
We leverage insights from the growing action space literature~\citep{czarnecki2018mix, farquhar2020growing} and consider a decoupled critic that increases its control resolution over the course of training. To this end, we define the discrete action sub-space at iteration $g$ as $\mathcal{A}^g \subset \mathcal{A}$ and modify the TD target to yield
\begin{align}
\label{eq:target_grow}
    y_t = \sum_{j=0}^{n-1}\gamma^{j}r\mleft(s_{t+j}, a_{t+j}\mright) + \gamma^n \sum_{j=1}^{M}  \max_{a^j_{t+1} \in \mathcal{A}^{g}} \frac{Q^j_{\theta^-}(\bm{s}_{t+n}, a^j_{t+n})}{M},
\end{align}
where $\epsilon$-greedy action sampling is analogously constrained to $\mathcal{A}^g$. The network architecture accommodates the full discretized action space from the start and constrains the active set via action masking, enabling masked action combinations to still profit from information propagation in the shared torso~\citep{van2017hybrid}. A schematic depiction of a decoupled agent with $5$-bin discretization and active $3$-bin subspace is provided in Figure~\ref{fig:gqn_schematic}.
In order to deploy such an agent we require a schedule for when to expand the active action space $\mathcal{A}^g \rightarrow \mathcal{A}^{g+1}$. Here, we consider two simple variations to limit engineering effort. First, we consider a linear schedule that doubles control resolution every $\frac{1}{N+1}$ of total training episodes, where $N$ indicates the number of subspaces $\mathcal{A}^g$. Second, we formulate an adaptive schedule based on an upper confidence bound inspired threshold over the moving average returns
\begin{align}
\label{eq:threshold}
    G_{\textbf{threshold}, t} = \mleft( 1.00 - 0.05\sgn\mu_{\textbf{MA}, t-1}^G \mright) \mu_{\textbf{MA}, t-1}^G + 0.90 \sigma_{\textbf{MA}, t-1}^G,
\end{align}
where $\mu_{\textbf{MA}}$ and $\sigma_{\textbf{MA}}$ are the moving average mean and standard deviation of the evaluation returns, respectively. The objective underestimates the mean by $5\%$ and expands the action space whenever the current mean return falls below the threshold $\mu_{t}^G<G_{\textbf{threshold}, t}$, signifying performance stagnation. This parameterization can avoid pre-mature expansion when exploring under sparse rewards, but alternative formulations are also applicable.
\begin{figure}[t!]
    \includegraphics[width=1.0\linewidth]{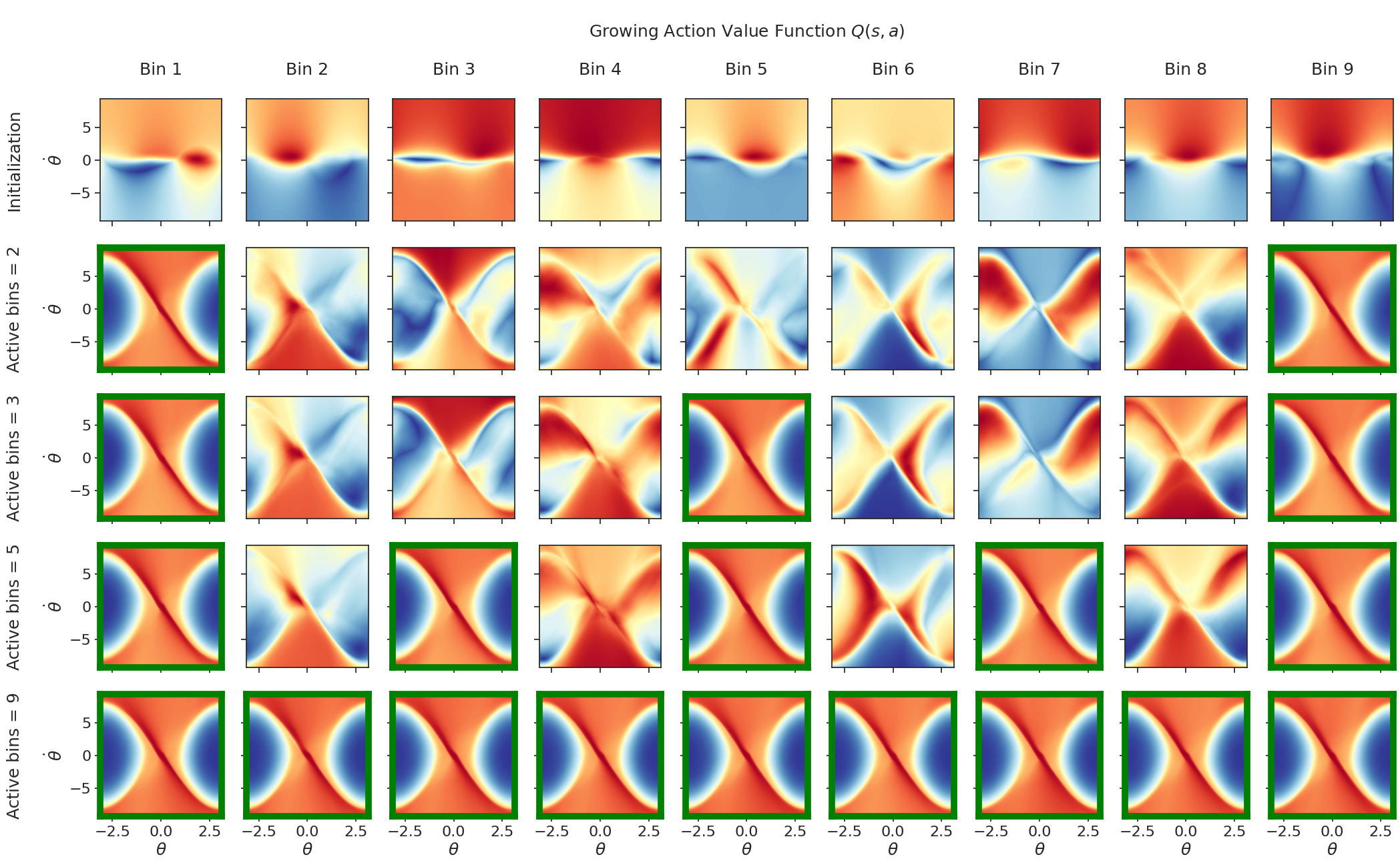}
\caption{State-action values for a pendulum swing-up task over the course of training (top to bottom). The active bins are outlined in green. The value predictions transition from random at initialization to structured upon activation. Inactive bins profit from the emergent structure within the shared network torso to warm-start their optimization.}
\label{fig:pendulum}
\end{figure}%
A qualitative example of our approach is provided in Figure~\ref{fig:pendulum}, where we visualize the state-action value function over the course of training on a pendulum swing-up task. We consider a GQN agent with discretization $2\rightarrow9$ (meaning $\{2, 3, 5, 9\}$) and provide learned values for each action bin starting at initialization and adding a row every time the action space is grown (top to bottom). The active bins are framed in green, where we observe accurate representation of the state-action value function for active bins, while the inactive bins still provide structured output due to high degree of weight sharing provided by our architecture.
In the following section, we provide quantitative results on a range of challenging continuous control tasks. We use the same set of hyperparameters throughout all experiments, unless otherwise indicated, following the general parameterization of \cite{seyde2022solving} with a simple multi-layer perceptron architecture and dimensionality $[512, 512]$. We evaluate mean performance with standard deviation across $4$ seeds and $10$ evaluation episode for each task.

\section{Experiments}
\begin{figure}[t!]
    \includegraphics[width=1.0\linewidth]{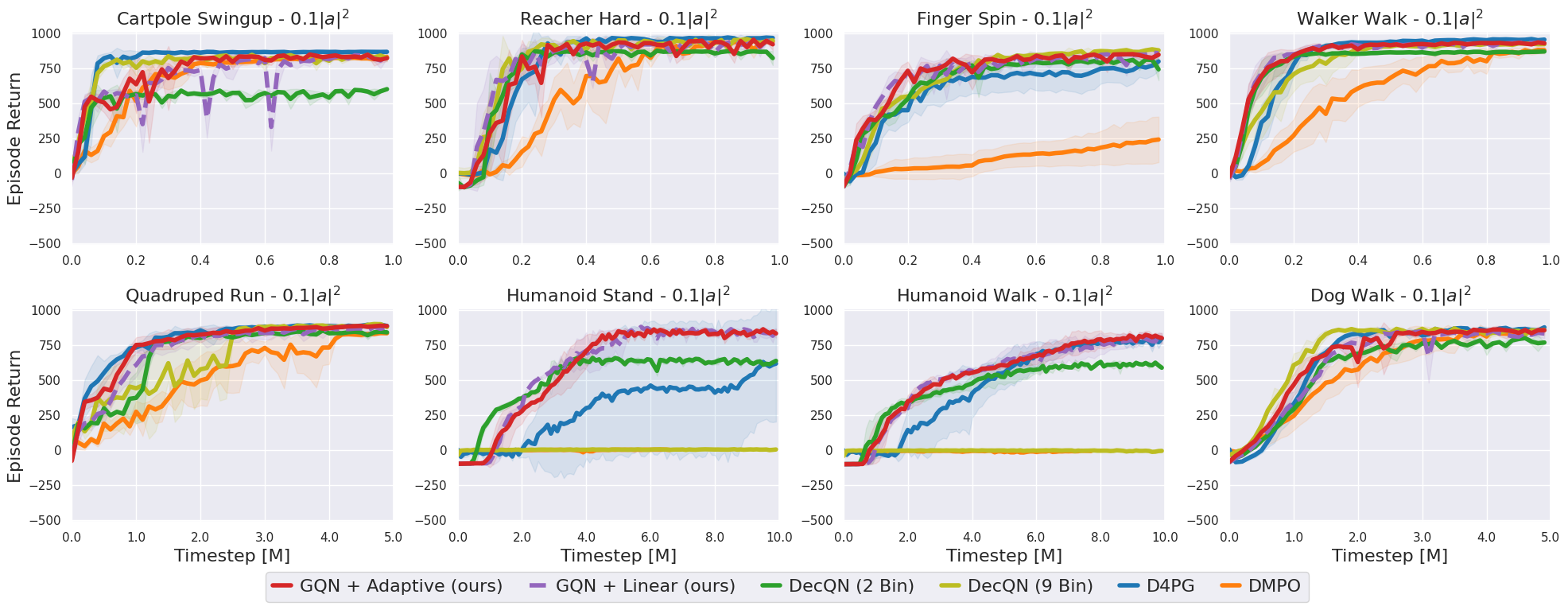}
\caption{Performance on tasks from the DeepMind Control Suite with action penalty $-0.1|a|^2$. Our GQN agent grows its action space from a $2$ bin to a $9$ bin discretization, where the linear and adaptive expansion schedules yield similar results. The GQN agent performs competitive to the discrete DecQN as well as the continuous D4PG and DMPO baselines, achieving noticeable improvements on the Humanoid Stand and Walk tasks.}
\label{fig:dmc01}
\vspace{-0.6cm}
\end{figure}

\begin{figure}[t!]
    \includegraphics[width=1.0\linewidth]{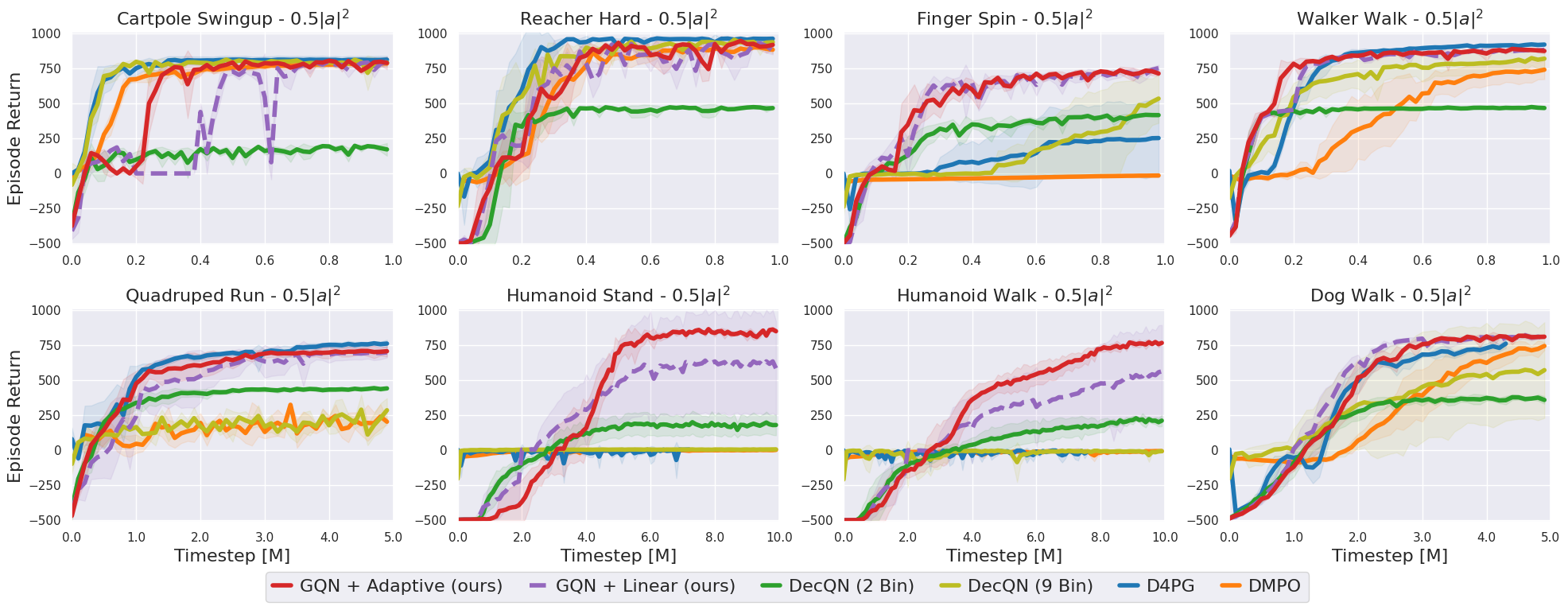}
\caption{Performance on tasks from the DeepMind Control Suite with action penalty $-0.5|a|^2$. Our GQN agent grows its action space from a $2$ bin to a $9$ bin discretization, where we observe benefits of the adaptive variant over the linear schedule. The GQN agent yields performance improvements over the discrete DecQN as well as the continuous D4PG and DMPO baselines, with particularly strong deltas on the Humanoid and Finger tasks.}
\label{fig:dmc05}
\vspace{-0.6cm}
\end{figure}

\begin{figure}[t!]
    \includegraphics[width=1.0\linewidth]{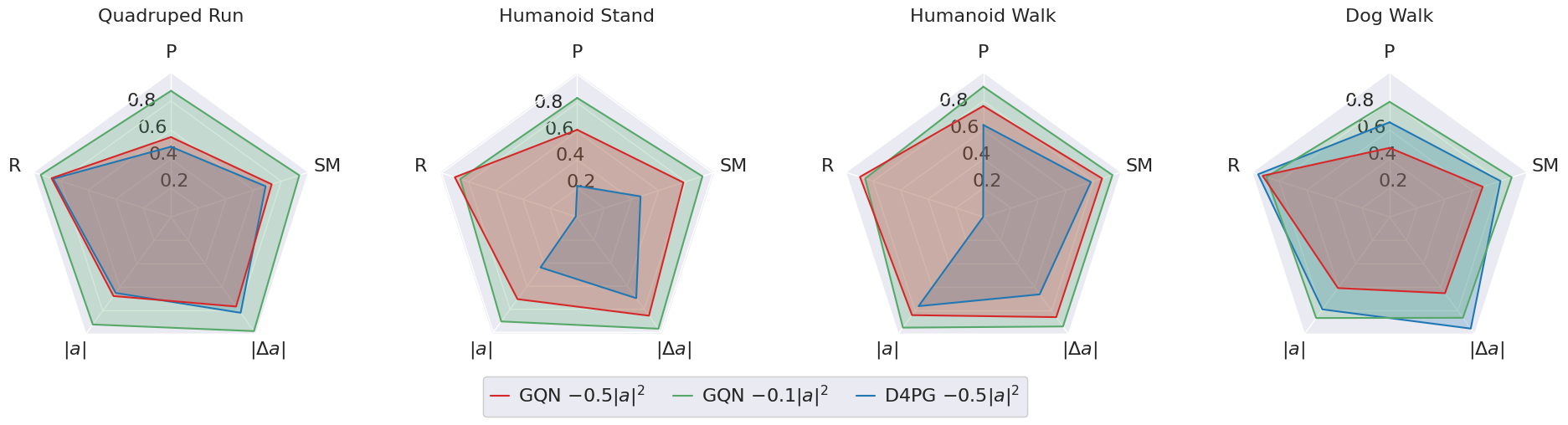}
\caption{Comparison of control smoothness and reward performance, relative to GQN without action penalties. Increasing the action penalty coefficient yields smoother control while only minor impact on the original task performance as measured by unconstrained reward $R$. The discrete GQN further improves upon the continuous D4PG agent.}
\label{fig:radar}
\vspace{-0.6cm}
\end{figure}

\begin{figure}[t!]
    \includegraphics[width=1.0\linewidth]{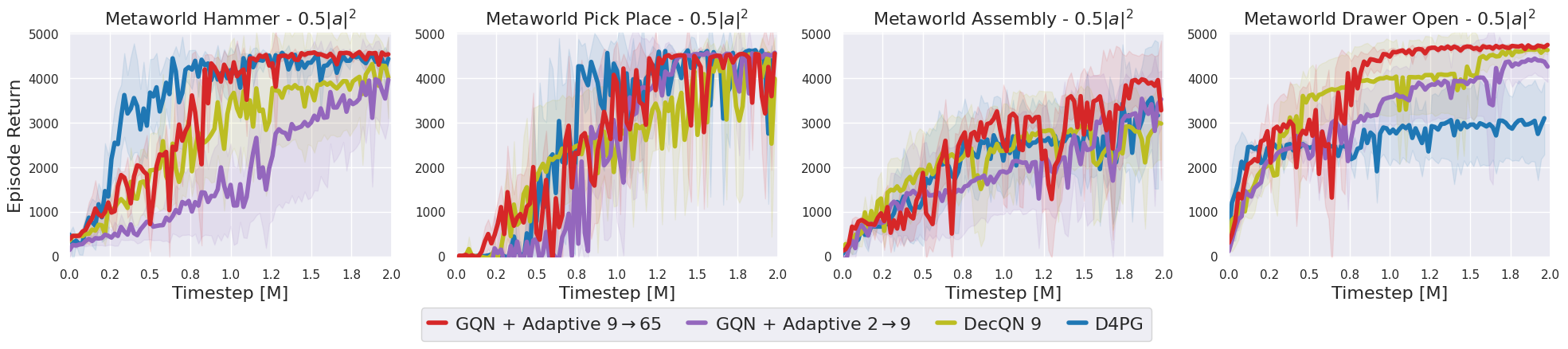}
\caption{Performance on manipulation tasks from MetaWorld with action penalty $-0.5|a|^2$. These tasks require control at the velocity level and are therefore more challenging to solve with extremely coarse discretization. We therefore investigate the scalability of our GQN agent and consider growing discretizations from $9$ up to $65$ bins. The resulting policy achieves stable learning and performs competitively with the continuous D4PG baseline while improving on the stationary $9$ bins DecQN agent.
}
\label{fig:meta}
\end{figure}

We evaluate our approach on a selection of tasks from the DeepMind Control Suite~\citep{tunyasuvunakool2020dm_control}, MetaWorld~\citep{yu2020meta}, and MyoSuite~\citep{MyoSuite2022}. The former two benchmarks generally do not consider action penalties and have previously been solved with bang-bang control~\citep{seyde2022solving}. We therefore focus on action-penalized task variations to encourage smooth control and highlight exploration challenges in the presence of penalty terms.
We first evaluate performance on tasks from the DeepMind Control Suite with action dimensionality up to $dim(\mathcal{A})=38$. We consider $2$ penalty weights $c_a \in \{0.1, 0.5\}$, such that rewards are computed as $r_t = r_t^o - c_a \sum_{j=1}^M {a_t^j}^2$ from original reward $r_t^o$.
We consider GQN agents that grow their action space discretization from $2$ to $9$ bins in each action dimension, where we evaluate both the linear and adaptive growing schedules discussed in Section~\ref{sec:gqn}. We compare performance against the state-of-the-art continuous control D4PG~\citep{barth2018distributed} and DMPO~\citep{abdolmaleki2018maximum} agents, while providing two discrete control DecQN agents with stationary action space discretization of $2$ or $9$ for reference.
The results in Figures~\ref{fig:dmc01} and~\ref{fig:dmc05} indicate the strong performance of GQN agents, with the adaptive schedule improving upon the linear schedule in terms of convergence rate and variance.
Growing control resolution further provides a clear advantage over the stationary DecQN agents both in terms of final performance (vs. DecQN 2) and exploration abilities (vs. DecQN 9). These observations mirror findings by~\cite{czarnecki2018mix}, where coarse control resolution was beneficial for early exploration, a characteristic that is amplified by the presence of action penalties.
We further observe the strong performance of discrete GQN agents compared to the continuous D4PG and DMPO agents.
In order to provide additional quantitative motivation for the presence of action penalties, we compare smoothness of the converged policies in Figure~\ref{fig:radar}. We consider the adaptive GQN agent with action penalties $c_a\in\{0.1, 0.5\}$ and the continuous D4PG agent with action penalty $c_a=0.5$. The metrics we consider are original non-penalized task performance, $R$, incurred action penalty, $P$, action magnitude, $|a|$, instantaneous action change, $|\Delta a|$, and the Fast Fourier Transform (FFT) based smoothness metric from~\cite{mysore2021regularizing}, SM. All metrics are normalized by the corresponding value achieved by the unconstrained GQN agent with $c_a=0.0$. The results indicate that increasing the action penalty yields noticeably smoother control signals while only having minor impact on the original task performance as measured by the unconstrained reward, $R$. We further find that smoothness of the discrete GQN agent is at least as good as for the continuous D4PG agent on the tasks considered (note that D4PG is unable to solve the Humanoid task variations, $R\approx0$).
We next extend our study to velocity-level control tasks for the Sawyer robot in MetaWorld. While acceleration-level control often provides sufficient filtering to interact favourably with highly discretized bang-bang exploration, velocity-level control tends to require more fine-grained inputs. We therefore investigate scalability of growing action spaces within decoupled Q-learning representations. To this end, we consider GQN agents with $2\rightarrow9$ and $9\rightarrow65$ (meaning $\{9, 17, 33, 65\}$) discretization as well as a stationary DecQN agent with $9$ bins. The results in Figure~\ref{fig:meta} indicate that initial bang-bang action selection is not well-suited for generating velocity-level actions, with the agent achieving good performance once transitioning to more fine-grained discretization (GQN $2\rightarrow9$). Interestingly, considering a larger growing action space with GQN $9\rightarrow65$ can surpass the performance of a stationary DecQN $9$ agent, despite the non-stationary optimization objective induced by the addition of finer action discretizations over the course of training. Performance of GQN $9\rightarrow65$ is furthermore competitive with the continuous D4PG agent on average.
Lastly, we stress-test our approach by considering a selection of tasks from the MyoSuite benchmark. The tasks require control of biomechanical models that aim to be physiologically accurate with $dim(\mathcal{A})=39$ and up to $dim(\mathcal{O})=115$, and should constrain applicability of simple decoupled Q-learning approaches such as GQN. Indeed, we find that the agent capacity becomes a limiting factor yielding overestimation errors that are further exacerbated by the large magnitude reward signals. We therefore extend the network capacity to $[512, 512]\rightarrow[2048, 2048]$ and lower the discount factor $\gamma=0.99\rightarrow0.95$ (alternatively, increasing multi-step returns $3\rightarrow5$ worked similarly well). With these parameter adjustments, we observe good performance as measured by task success at the final step of an episode. This further underlines the surprising effectiveness that decoupled discrete control can yield in continuous control settings and the benefit of adaptive control resolution change over the course of training.
\begin{figure}[t!]
    \includegraphics[width=1.0\linewidth]{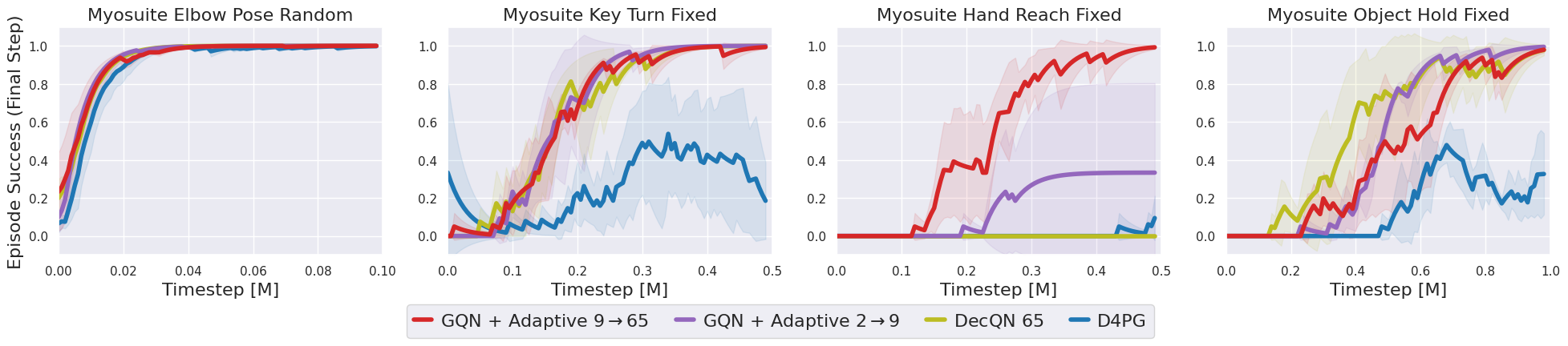}
\caption{Performance for controlling biomechanical models from the MyoSuite as measured by task success at termination. These continuous control tasks stress test growing decoupled discrete action spaces, due to their dimensionality and inherent complexity. Increasing the network capacity and adjusting the discount factor to mitigate overestimation, we observe strong performance for growing action spaces up to a discretization of $65$ bins.}
\label{fig:myo}
\end{figure}%

\section{Conclusion}
In this work, we investigate the application of growing action spaces within the context of decoupled Q-learning to efficiently solve continuous control tasks. Our Growing Q-Networks (GQN) agent leverages a linear value decomposition along actuators to retain scalability in high-dimensional action spaces and adaptively increases control resolution over the course of training. This enables coarse exploration early during training without reduced control smoothness and accuracy at convergence. The resulting agent is robust and performs well even for very fine control resolutions despite inherent non-smoothness in the optimization objective arising at the transition between resolution levels. While GQN as a critic-only method displays very strong performance compared to recent continuous actor-critic methods on the tasks considered, we also investigate scenarios that prove challenging for decoupled discrete controllers as exemplified by velocity-level control of simulated manipulators or applications to control of biomechanical models. Interesting avenues for future work include addressing coordination challenges in increasingly high-dimensional action spaces and mitigating overestimation bias. Generally, GQN provides a simple yet capable agent that efficiently bridges the gap between between coarse exploration and solution smoothness through adaptive control resolution refinement.

\acks{Tim Seyde, Peter Werner, Wilko Schwarting, and Daniela Rus were supported in part by the Office of Naval Research (ONR) Grant N00014-18-1-2830, Qualcomm, and the United States Air Force Research Laboratory and the Department of the Air Force Artificial Intelligence Accelerator under Cooperative Agreement Number FA8750-19-2-1000. The views and conclusions contained in this document are those of the authors and should not be interpreted as representing the official policies, either expressed or implied, of the Department of the Air Force or the U.S. Government. The U.S. Government is authorized to reproduce and distribute reprints for Government purposes notwithstanding any copyright notation herein. The authors further would like to acknowledge the MIT SuperCloud and Lincoln Laboratory Supercomputing Center for providing HPC resources.}

\bibliography{bibliography}

\end{document}